\documentclass[10pt,letterpaper]{article}
\usepackage[margin=0.75in,columnsep=0.25in]{geometry}
\usepackage[T1]{fontenc}
\usepackage{lmodern,microtype}
\usepackage{amsmath,amssymb,booktabs,array,graphicx,tabularx}
\usepackage[numbers,sort&compress]{natbib}
\usepackage[font=small,labelfont=it]{caption}
\usepackage{xurl}
\usepackage{titlesec}
\titleformat{\section}{\normalfont\fontsize{12}{14}\selectfont\bfseries\raggedright}{\thesection}{0.65em}{}
\titleformat{\subsection}{\normalfont\normalsize\bfseries\raggedright}{\thesubsection}{0.65em}{}
\titlespacing*{\section}{0pt}{2.2ex plus .5ex minus .2ex}{1ex plus .2ex}
\titlespacing*{\subsection}{0pt}{1.7ex plus .4ex minus .2ex}{.7ex plus .2ex}
\makeatletter
\renewcommand{\@maketitle}{%
  \newpage
  \begin{center}%
    \hrule height 2pt
    \vskip 10pt
    {\fontsize{20}{24}\selectfont\bfseries \@title\par}%
    \vskip 10pt
    \hrule height .5pt
    \vskip 12pt
    {\large \@author\par}%
  \end{center}%
  \vskip 12pt
}
\makeatother

\makeatletter
\newcommand{\makepaperfrontmatter}[1]{%
  \twocolumn[{%
    \@maketitle
    \begin{minipage}{\textwidth}%
      \normalfont\normalsize
      \hrule height .4pt
      \vspace{7pt}%
      \noindent\textbf{Abstract}---#1\par
      \vspace{7pt}%
      \hrule height .4pt
    \end{minipage}%
    \vspace{16pt}%
  }]%
}
\makeatother

\usepackage[hidelinks]{hyperref}
\usepackage{bookmark}
\usepackage{tikz}
\usepackage{pgfplots}
\usepgfplotslibrary{groupplots,fillbetween}
\pgfplotsset{compat=1.18}
\definecolor{arbHuman}{HTML}{9A6DD7}
\definecolor{arbGreen}{HTML}{28B487}
\definecolor{arbBlue}{HTML}{348CE5}
\definecolor{arbOrange}{HTML}{F39B35}
\definecolor{arbCoral}{HTML}{ED6A5A}
\definecolor{arbLime}{HTML}{9CCB4B}
\definecolor{arbGold}{HTML}{E6BD33}
\definecolor{arbRose}{HTML}{E66BA6}
\definecolor{arbIndigo}{HTML}{6D7CE8}
\definecolor{arbClay}{HTML}{D98C62}
\definecolor{arbOrchid}{HTML}{C783DD}
\definecolor{arbCyan}{HTML}{35BFD0}
\definecolor{arbInk}{HTML}{343837}
\definecolor{arbGrid}{HTML}{E4DED6}
\newcommand{\arbMarkerSize}{1.5pt}
\newcommand{\arbHighlightSize}{3pt}
\newcommand{\arbDataLine}{.8pt}
\newcommand{\arbGuideLine}{.5pt}
\newcommand{\arbMarkerEdge}{.25pt}
\tikzset{
  arborea data/.style={line width=\arbDataLine},
  arborea guide/.style={line width=\arbGuideLine},
  arborea point/.style={mark size=\arbMarkerSize},
  arborea highlight/.style={mark size=\arbHighlightSize},
  arborea marker edge/.style={line width=\arbMarkerEdge},
}
\pgfplotsset{arborea legend below/.style={
  legend style={at={(.5,0)},anchor=north,yshift=-.54in},
},arborea density/.style={
  no marks,arborea data,fill opacity=.25,draw opacity=1,area legend,
},arborea axis/.style={
  scale only axis,width=2.25in,height=1.125in,
  font=\footnotesize,tick label style={font=\footnotesize\bfseries\boldmath,inner sep=.3333em},
  title style={font=\footnotesize\bfseries\boldmath},
  label style={font=\footnotesize\bfseries\boldmath,inner sep=.3333em},
  legend style={font=\scriptsize,draw=none,fill=none,cells={anchor=west}},
  axis lines=left,axis line style={-},tick align=outside,
  axis line style={line width=.6pt,draw=arbInk},
  grid=major,grid style={arbGrid,line width=.3pt},
  minor tick num=0,major tick length=2pt,
  scaled ticks=false,enlargelimits=false,
  clip marker paths=true,
  every axis plot/.append style={arborea data,arborea point,
    mark options={arborea marker edge}},
},arborea scatter/.style={
  arborea axis,height=2.25in,
}}

\usetikzlibrary{arrows.meta,positioning}
\definecolor{arbYellow}{HTML}{EDA100}
\tikzset{visualbox/.style={draw=black!45, rounded corners=1pt, align=center,
  font=\footnotesize, inner sep=4pt, minimum height=10mm},
  visualarrow/.style={->, draw=black!65, line width=.6pt}}

\newlength{\readerfigurewidth}
\makeatletter
\newcommand{\recordfigurewidth}{%
  \ifdim\dimexpr\pgf@picmaxx-\pgf@picminx\relax>\readerfigurewidth
    \global\readerfigurewidth=\dimexpr\pgf@picmaxx-\pgf@picminx\relax
  \fi}
\makeatother
\newenvironment{readerfigure}{%
  \begin{figure}[!htbp]\centering
  \global\readerfigurewidth=0pt
  \tikzset{every picture/.append style={execute at end picture={\recordfigurewidth}}}%
  \captionsetup{type=figure}%
  \let\readeroriginalcaption\caption
  \renewcommand{\caption}{%
    \ifdim\readerfigurewidth>0pt
      \captionsetup{width=\readerfigurewidth}%
    \fi
    \readeroriginalcaption}%
}{\end{figure}}
\newenvironment{readerplot}{%
  \begin{readerfigure}
  \tikzset{every picture/.append style={trim axis left,trim axis right}}%
}{\end{readerfigure}}
\hypersetup{pdftitle={Concept-Guided Prompting Strategies Increase Diversity in LLM-Generated Recipe Populations},pdfauthor={Stuart Farmer}}
\graphicspath{{figures/}{figure_review/figures/}}

\title{Concept-Guided Prompting Strategies Increase Diversity in LLM-Generated Recipe Populations}

\author{
  \textbf{Stuart Farmer} \\
  \href{https://casadilauro.com}{Casa Dilauro Independent Research House} \\
  \href{mailto:stuart@casadilauro.com}{\texttt{stuart@casadilauro.com}} \\
  \url{https://casadilauro.com} \\
  \url{https://github.com/casadilauro/concept-guidance}
}
\date{}

\makeatletter
\renewcommand{\makepaperfrontmatter}[1]{%
  \par\begingroup
  \normalfont\fontsize{10}{11}\selectfont
  \setlength{\parindent}{0pt}
  \setlength{\parskip}{5.5pt}
  \noindent\begin{minipage}{\textwidth}
    \vbox{%
      \hsize\textwidth\linewidth\hsize
      \vskip 0.1in
      \hrule height 4pt
      \vskip 0.25in\vskip -\parskip
      \centering
      {\fontsize{17}{20}\selectfont\bfseries\@title\par}
      \vskip 0.29in\vskip -\parskip
      \hrule height 1pt
      \vskip 0.09in
      \begin{tabular}[t]{c}\rule{0pt}{24pt}\@author\end{tabular}\par
      \vskip 0.3in minus 0.1in
    }
    \vskip 0.075in
    \hrule height .4pt\vspace{7pt}
    \centerline{\fontsize{12}{14}\selectfont\bfseries Abstract}
    \vspace{0.5ex}
    \setlength{\leftmargini}{3pc}
    \begin{quote}#1\par\end{quote}
    \vspace{7pt}\hrule height .4pt
  \end{minipage}\par
  \endgroup\vspace{16pt}}
\makeatother
\begin{document}
\raggedbottom
\renewcommand{\floatpagefraction}{.8}
\twocolumn[{\makepaperfrontmatter{
Novelty in solution generation supports automated discovery, but LLM outputs can remain repetitive. To address this, we propose a simple concept-guided prompting method that generates thematic concepts before responses within each of these concepts. To demonstrate, we generated 1,000 recipes for a recipe book task and compared the results against a human-authored corpus. Two naive prompting strategies were compared as baselines. Concept-guided prompting outperformed both baselines in all cases and reached semantic coverage of 47.0\% versus 35.1\% and 13.1\% for the prior-title and nonce prompting strategies, respectively. These findings suggest broader population exploration and diversity, but remain contained to this specific task. Further work is needed to determine the quality of solutions and generalizability.
}}]

\section{Introduction}

Large language models have been shown to be effective components of self-improvement and optimization systems across many problem domains. These systems generally query an LLM for a solution to a predetermined problem, score the proposed solution using a standardized evaluator, and return the score to the LLM as feedback to help generate the next population of solutions.

As is the case with most optimizers, overfitting and getting stuck in local minima are problems for LLM-guided search. Some systems, such as FunSearch, attempt to address this by using tactics inspired by evolutionary computing such as sampling across an entire population to mutate and combine ideas together \citep{romeraparedes2024funsearch}.

Nonetheless, search can prematurely converge on narrow regions even when these tactics are employed. Using isolated island populations that search independently and in parallel may help mitigate this, as is the case in the ShinkaEvolve algorithm, another LLM-guided search system. However, this tactic does not explicitly measure coverage of the behavior space \citep{lange2025shinkaevolve}.

Advances in other domains have attempted to solve this issue. Quality-diversity optimization is a class of evolutionary algorithms that considers novelty alongside raw performance against an objective function \citep{cully2017qdframework}. One such algorithm, MAP-Elites, divides the space of all possible solutions into behavioral cells and retains the highest-performing solution for each of these cells. This helps uncover multiple approaches to performing a task \citep{mouret2015mapelites}. Applications that utilize these methods include robotic behavior repertoires \citep{cully2015robots} and procedural game content generation: MAP-Elites variants have produced playable Mario scenes with varied gameplay characteristics \citep{fontaine2021mario}.

\begin{readerfigure}
\includegraphics[width=\linewidth]{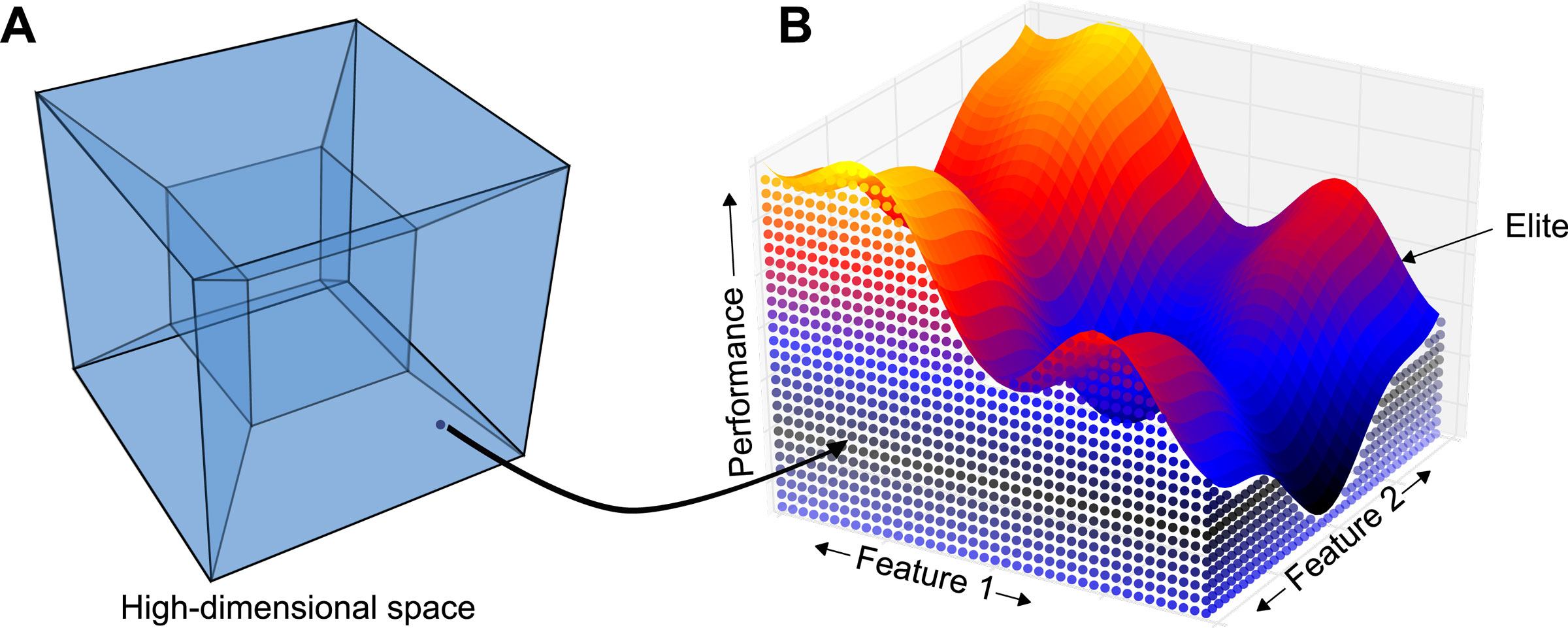}
\caption{MAP-Elites maps a high-dimensional search space (A) onto a feature
space (B), retaining the best-performing solution found in each region. Height
and color represent performance across two feature dimensions. Reproduced
from Mouret and Clune~\citep{mouret2015mapelites}, Figure~1 (arXiv:1504.04909v1).
This is a conceptual illustration, not a result from the recipe experiment.}
\label{fig:map-elites-concept}
\end{readerfigure}

The combination of the two, LLM-guided search with quality-diversity methods, has been proposed in a method known as Quality-Diversity through AI Feedback (QDAIF). QDAIF uses an LLM both to generate the solution and to evaluate its quality and diversity \citep{bradley2023qdaif}.

However, this method requires additional LLM calls and inference beyond candidate generation, which can increase search time and cost. This motivates research into approaches that promote diversity during generation itself \citep{madaan2023selfrefine,wang2024implicit}.

Thus, we ask a simple question: can a prompting structure alone encourage broader populations without an iterative feedback process?

The solution we came up with we call concept-guided prompting, which decomposes a problem into a higher-level layer of thematic concepts that guide solution generation. These concepts are model-generated themes supplied as context for subsequent requests.

\begin{readerfigure}
\begin{tikzpicture}[
  point/.style={circle,draw=black!60,fill=white,minimum size=4mm,inner sep=0pt},
  edge/.style={draw=black!55,line width=.6pt},
  level label/.style={anchor=east,font=\sffamily\footnotesize}]
\definecolor{concept1}{HTML}{929CDD}
\definecolor{concept2}{HTML}{FEFFCC}
\definecolor{concept3}{HTML}{F6AFB4}
\definecolor{solution11}{HTML}{0C1878}
\definecolor{solution12}{HTML}{2A3789}
\definecolor{solution13}{HTML}{046664}
\definecolor{solution21}{HTML}{6CC324}
\definecolor{solution22}{HTML}{FCFE04}
\definecolor{solution23}{HTML}{FC9A04}
\definecolor{solution31}{HTML}{E73910}
\definecolor{solution32}{HTML}{CA1622}
\definecolor{solution33}{HTML}{9C0264}
\node[level label] at (-0.3,0) {Problem Context};
\node[level label] at (-0.3,-0.9) {Concepts};
\node[level label] at (-0.3,-1.8) {Guided Solutions};
\node[point] (root) at (2.3,0) {};
\foreach \x/\id in {0.7/1,2.3/2,3.9/3} {
  \node[point,fill=concept\id] (concept\id) at (\x,-0.9) {};
  \draw[edge] (root) -- (concept\id);
  \foreach \offset/\child in {-0.5/1,0/2,0.5/3} {
    \node[point,fill=solution\id\child] (solution\id\child) at (\x+\offset,-1.8) {};
    \draw[edge] (concept\id) -- (solution\id\child);
  }
}
\end{tikzpicture}

\caption{A tree of height two connects one problem context to three concepts,
each guiding three solutions.}
\label{fig:concept-tree}
\end{readerfigure}

The idea is that the semantic taxonomy would add additional information and guidance to the subsequent solution sets, as each smaller set would have a bit of further steering toward a specific behavioral space.

For example, rather than generating 1,000 solutions and hoping they would differ from one another, we could generate 100 concepts and then 10 solutions under each concept. Ideally, this would allow us to only need those 100 concepts to be semantically diverse, as the solutions under each concept would be related to each other more closely, but maintain the same breadth of behavioral space that the higher-level concepts have.

\section{Experimental design}

\label{sec:experiment-design}

To test this hypothesis, we used a toy task: generating recipes for a hypothetical recipe book. The generation prompts requested three components: a title, an ingredient list, and preparation instructions (Table~\ref{tab:recipe-model}). No separate description field was requested. These components were stored together as recipe text, alongside the extracted title and collection or generation metadata.

\begin{table}[!htbp]
\captionsetup{type=table}
\caption{Requested recipe content. All three components form one plain-text recipe.}
\label{tab:recipe-model}
\small
\begin{tabularx}{\linewidth}{@{}l >{\raggedright\arraybackslash}X@{}}
\toprule
Component & Content \\
\midrule
Title & A single line naming the recipe. \\
Ingredients & A list of ingredients, including quantities where appropriate. \\
Instructions & An ordered sequence of preparation steps. \\
\bottomrule
\end{tabularx}
\end{table}

Our control group was a human-authored corpus of 168 recipes downloaded from Text Only Recipes. The original page URLs and complete recipe texts were retained; human recipes kept their source formatting. Each complete recipe text was encoded as one 3,072-dimensional vector using \path{text-embedding-3-large} through OpenRouter. The same encoder was used for all populations.

A large embedding model was used to convert unstructured text into standardized vectors which we could then run comparative analyses on. This was based on the idea that learned text embeddings can support semantic similarity comparisons using cosine similarity \citep{reimers2019sentencebert}.

Thus, embedded recipes that are closer together should be more semantically similar. This is used as a proxy, as it is not a direct measure of culinary similarity or recipe quality.

The experiment is then as follows: propose two simpler, baseline prompting strategies alongside our concept-guided method. Have each strategy generate 1,000 recipes for its own corpus. Afterward, we can analyze them internally, against each other, and against the human corpus to understand the semantic and geometric differences to better understand what semantic space was produced in total by each strategy.

\section{Prompting Strategies}

\subsection{Nonces}

The first baseline strategy is to just repeatedly ask the LLM to generate a recipe. To avoid absolute determinism, a small nonce is inserted into the prompt. This helps activate differing neurons in the LLM and explore the intrinsic, non-contextual space of the model itself.

\begin{readerfigure}
\tikzset{visualbox/.append style={minimum height=7mm}}
\begin{tikzpicture}[>=Latex, x=1cm, y=1cm]
\foreach \y/\i [count=\row] in {0/1,-1/2,-2.6/{1{,}000}} {
  \node[visualbox, text width=25mm, fill=arbOrange!12] (p\row) at (1.5,\y)
    {Recipe prompt\\+ nonce $\i$};
  \node[visualbox, text width=9mm] (l\row) at (4.4,\y) {LLM};
  \node[visualbox, text width=18mm] (r\row) at (6.9,\y) {Recipe $\i$};
  \draw[visualarrow] (p\row)--(l\row);
  \draw[visualarrow] (l\row)--(r\row);
}
\foreach \x in {1.5,4.4,6.9}
  \node[font=\footnotesize] at (\x,-1.72) {$\vdots$};
\end{tikzpicture}
\caption{Independent requests with different nonces.}
\label{fig:prompting-nonce}
\end{readerfigure}

\subsection{Titles}

The second baseline strategy is to generate recipes sequentially. After each new recipe, the title of that recipe is appended to a list of all previously generated recipe titles. That list is included in the prompt for the LLM with explicit instructions not to return recipes that have already been generated. This strategy was designed to measure what a user may initially think to do when requiring diversity from an LLM and therefore serves as a useful baseline to see if our concept-guided method is actually better than a basic strategy apart from the simple isolated prodding of an LLM for recipes out of the blue.

\begin{readerfigure}
\tikzset{visualbox/.append style={minimum height=7mm}}
\begin{tikzpicture}[>=Latex, x=1cm, y=1cm]
\node[visualbox, text width=27mm, fill=arbBlue!10] (p) at (1.6,0)
  {Recipe prompt\\+ previous titles\\``Avoid these recipes''};
\node[visualbox, text width=9mm] (l) at (4.5,0) {LLM};
\node[visualbox, text width=18mm] (r) at (7,0) {New recipe};
\node[visualbox, text width=27mm, fill=arbBlue!10] (t) at (4.5,-1.5)
  {Append its title\\to the title list};
\draw[visualarrow] (p)--(l);
\draw[visualarrow] (l)--(r);
\draw[visualarrow] (r.south) |- (t.east);
\draw[visualarrow] (t.west) -| (p.south);
\end{tikzpicture}
\caption{Sequential requests, starting with an empty title list.}
\label{fig:prompting-titles}
\end{readerfigure}

\subsection{Concepts}

The third strategy is, of course, our proposed strategy. At first, the concept layer is generated in a batch so that the LLM does not repeat concepts. After that, each of the concepts is injected into a prompt that tells the LLM to generate a set of recipes related to the provided concept.

\begin{readerfigure}
\tikzset{visualbox/.append style={minimum height=7mm}}
\begin{tikzpicture}[>=Latex, x=1cm, y=1cm]
\node[visualbox, text width=34mm, fill=arbGreen!10] (p) at (4,0)
  {Generate 100 concepts};
\node[visualbox, text width=24mm] (l) at (4,-.95) {LLM: one batch};
\draw[visualarrow] (p)--(l);
\foreach \x/\i in {1.3/1,4/2,6.7/100} {
  \node[visualbox, text width=20mm, fill=arbGreen!10] (c\i) at (\x,-2.1)
    {Concept $\i$};
  \node[visualbox, text width=20mm] (g\i) at (\x,-3.2)
    {LLM\\recipe batch};
  \node[visualbox, text width=20mm] (r\i) at (\x,-4.3)
    {10 recipes};
  \draw[visualarrow] (l.south)--(c\i.north);
  \draw[visualarrow] (c\i)--(g\i);
  \draw[visualarrow] (g\i)--(r\i);
}
\node[font=\footnotesize, inner sep=0pt] at (5.35,-2.1) {$\cdots$};
\end{tikzpicture}
\caption{Concept layer first, then solution requests.}
\label{fig:prompting-concepts}
\end{readerfigure}

\subsection{Model}

The \path{openai/gpt-oss-120b:nitro} model from OpenRouter was used for generation. Temperature and top-p were set to 0 and 1, respectively, to maximize the determinism of the model so that the strategies were what provided diversity, not probabilistic sampling of the output vector of the large language model.

\subsection{Execution}

The experiment, as mentioned earlier, had each of the 3 strategies generate 1,000 recipes in our desired format. Requests were made over HTTP to the OpenRouter server according to the maximum concurrency ability of each strategy. The nonce strategy was completely parallel. The title strategy was serial. The concept strategy first had to generate the concepts, and then after that it was parallel across each of the concepts.

The concept strategy first generated 100 concepts and then 10 recipes under each concept for a total of 1,000 recipes, adhering to the protocol.

Table~\ref{tab:design} records the request counts, response limits, and concurrency.

\begin{table*}[t]
\centering\small
\caption{Execution settings. All strategies used the same model, temperature
and top-$p$, and retained 1,000 recipes. Token caps are per-response limits,
not observed token consumption or evidence of equal computational budgets.}
\label{tab:design}
\renewcommand{\arraystretch}{1.15}
\begin{tabularx}{\textwidth}{@{}>{\raggedright\arraybackslash}Xrrr>{\raggedright\arraybackslash}X@{}}\toprule
Strategy & Calls & Recipes/call & Token cap & Execution \\\midrule
Nonce & 1,000 & 1 & 1,200 & Up to four workers \\
Prior titles & 1,000 & 1 & 1,200 & Sequential \\
Concept-guided & $101+3$ & 10$^\dagger$ & 16,000 & Up to four workers for context-level requests \\
\bottomrule\end{tabularx}
\par\smallskip
\parbox{\textwidth}{\footnotesize $^\dagger$Ten recipes per main recipe batch.
The concept-guided strategy first generated all 100 contexts in one call, followed by 100
recipe-batch calls. Three additional repair calls recovered 19 omitted texts.}
\end{table*}

Each request carried a deterministic SHA-256 nonce derived from its position to support identification and resumption. This identifier did not ensure deterministic model responses, nor was it meant to.

\subsection{Considerations}

During the execution of the protocol, the concept-guided strategy required three targeted repair calls to recover 19 omitted recipe texts. These calls requested missing items rather than evaluating or improving recipe quality. Repair responses were retained separately and traced to the recovered texts.

A parser kept the first occurrence of each identifier and requested missing identifiers in smaller batches, allowing up to three repair attempts before failure. Because saved rows can retain parent-request metadata, repaired texts were linked to their actual repair responses during the audit.

All stored texts were retained, including incomplete responses. The prior-title population contained 78 length-limited responses, including 17 recorded format failures; neither of the other generated populations had recorded failures of these types. These diagnostics describe output format and completeness, not culinary validity. Differences in response limits and batching remained part of the comparison, and request counts alone did not establish cost or latency efficiency.

These methods and all else not covered in this paper are available for inspection and evaluation in the GitHub repository at \href{https://github.com/CasaDilauro/concept-guidance}{\nolinkurl{github.com/CasaDilauro/concept-guidance}}.

\section{Results}

Following the protocol, the three generated recipe populations and existing human-authored corpus were processed into embedding vectors and analyzed. A total of 7 analyses were conducted to form a multifaceted and holistic view of what was generated and how diverse each set actually was on a semantic level without having to score recipes manually or rely on qualitative methods.

\subsection{Internal Local Spacing}
\label{sec:local-spacing}

The first analysis was made to understand how diverse each population was internally. In other words, how different were the recipes generated from each other in the same population?

To determine this, we used a local spacing measure. Coordinate-wise min-max scaling to [0, 3] across all four populations was applied jointly. Then, for each scaled embedding, we found the k nearest other recipes in that population, excluding the recipe itself.

The average of the squared Euclidean distances for each sample was taken according to:
\begin{equation}
\rho_i(k)=\frac{1}{k}\sum_{j\in N_k(i)}\lVert z_i-z_j\rVert^2.
\end{equation}
A mean of all these distances, $\bar{\rho}(k)=n^{-1}\sum_{i=1}^{n}\rho_i(k)$, where $n$ is the number of recipes in that population, provides the population's internal local spacing. Our main analysis uses $k=15$.

\begin{readerplot}
\pgfplotsset{title={(a) Local recipe neighborhoods}}
\input{figures/tikz/results_local_spacing}\par
\renewcommand{\thefigure}{\arabic{figure}a}
\caption{Local neighborhoods in the measured recipe populations. Within each population, one circled recipe nearest the median spacing score is linked to its 15 nearest other recipes.}
\label{fig:local-spacing}
\end{readerplot}

\begin{readerplot}
\ContinuedFloat
\pgfplotsset{title={(b) Local spacing distribution}}
% Generated by analysis/export_results_method_plots.py; measured recipe data.
\begin{tikzpicture}
\begin{axis}[arborea axis,xmin=0,xmax=2000.0,ymin=0,ymax=0.026,xlabel={Mean squared neighbor distance, $\rho_i(15)$},ylabel={Density},legend columns=2,arborea legend below]
\addplot[arbBlue,dashed,arborea density,fill=arbBlue] table[x=x,y=y] {figures/tikz/data/results_spacing_kde_prior_titles.dat}\closedcycle;
\addlegendentry{Prior titles}
\addplot[arbBlue,dashed,no marks,forget plot] coordinates {(354.30894899350096,0) (354.30894899350096,0.026)};
\addplot[arbOrange,densely dotted,arborea density,fill=arbOrange] table[x=x,y=y] {figures/tikz/data/results_spacing_kde_nonce.dat}\closedcycle;
\addlegendentry{Nonce}
\addplot[arbOrange,dashed,no marks,forget plot] coordinates {(37.027580963747084,0) (37.027580963747084,0.026)};
\addplot[arbGreen,dashdotted,arborea density,fill=arbGreen] table[x=x,y=y] {figures/tikz/data/results_spacing_kde_hierarchical.dat}\closedcycle;
\addlegendentry{Concept-guided}
\addplot[arbGreen,dashed,no marks,forget plot] coordinates {(562.1167044209221,0) (562.1167044209221,0.026)};
\addplot[arbHuman,solid,arborea density,fill=arbHuman] table[x=x,y=y] {figures/tikz/data/results_spacing_kde_baseline.dat}\closedcycle;
\addlegendentry{Human}
\addplot[arbHuman,dashed,no marks,forget plot] coordinates {(1055.4866503392464,0) (1055.4866503392464,0.026)};
\end{axis}
\end{tikzpicture}\par
\renewcommand{\thefigure}{\arabic{figure}b}
\caption{Gaussian kernel density estimates of local spacing scores, with
Scott's bandwidth per population and reflection at zero. Each curve integrates
to one; vertical dashed lines show observed means.}
\label{fig:local-spacing-histogram}
\end{readerplot}

Larger values of local spacing indicate greater average local separation in the scaled embedding space, which we interpret as greater local semantic variation. As mentioned earlier, embeddings offer us a proxy, and as such this measure is a geometric proxy as well and does not measure the quality of the responses.

Embeddings were scaled together across all populations, which keeps the coordinates comparable, but does not equalize population sizes. Fixed-$k$ spacing can depend on the number of recipes.

The measurement uses the full embeddings, not a two-dimensional projection.

\begin{table}[!htbp]
\captionsetup{type=table}
\caption{Mean local spacing in squared-distance units after joint coordinate-wise scaling. The human value provides context; its smaller population prevents an equal-density comparison with the generated populations. Across the results tables, italics identify the human control and bold marks the best generated-strategy value for each measure.}
\label{tab:local-spacing}
\small\setlength{\tabcolsep}{3pt}
\begin{tabularx}{\linewidth}{@{}Xr@{}}\toprule
Population & \shortstack{Mean spacing\\$\bar{\rho}(15)$} \\\midrule
\textit{Human reference} & \textit{1055.5} \\
Concept-guided & \textbf{562.1} \\
Prior titles & 354.3 \\
Nonce & 37.0 \\
\bottomrule\end{tabularx}
\end{table}

Among the three equally sized generated populations, concept-guided prompting had the highest mean local spacing: 1.59 times the prior-title value and 15.18 times the nonce value. Thus, even their nearest neighbors were farther away on average for concept-guided recipes, supporting greater local semantic variation within this saved population.

The density curves show how this difference is distributed across recipes. Nonce scores are concentrated near the low end, indicating closely packed local neighborhoods. Prior-title scores occupy an intermediate range, while the concept-guided distribution is shifted toward larger values. The distinction is therefore visible across the distributions as well as in their means.

The human reference had the highest observed mean spacing, but the unequal population sizes limit its interpretation. These results support greater internal local variation for concept-guided prompting relative to the two generated baselines; they do not establish similarity to human recipes or variation across independent generation runs.

\subsection{Internal Participation Ratio}
\label{sec:participation-ratio}

Variation along each dimension of the embedding vectors could indicate the amount of geometric reach a population covers. However, if that variation occurs on only a few dimensions, it could be a false positive signal of a sufficiently diverse population.

Thus, the participation ratio was calculated across each population to understand how well each strategy captured true variation in the semantic geometry of its generated population.

For this calculation, each recipe's embedding vector was first normalized to unit length. Within each sample, we then subtracted the average vector, centering the population around zero. We calculated the covariance eigenvalues, $\lambda_i$, which measure variation along the population's principal directions. These directions are combinations of the original embedding coordinates, ordered from greatest to least variance.

The participation ratio summarizes how evenly variation is distributed across these directions:
\begin{equation}
\begin{aligned}
\mathrm{PR}&=\frac{(\sum_i\lambda_i)^2}{\sum_i\lambda_i^2}=\frac{1}{\sum_i p_i^2},\\
p_i&=\frac{\lambda_i}{\sum_j\lambda_j}.
\end{aligned}
\end{equation}
Here, $p_i$ is the proportion of total variance carried by direction $i$. If all variation occurs along one direction, PR equals 1. If variation is distributed equally across ten directions, PR equals 10. Unequal contributions produce intermediate values.

Because sample size limits the number of observable directions, comparisons used matched samples of 168 recipes. The human population was evaluated once, while each generated population was evaluated across 50 random subsets of 168 recipes, sampled without replacement within each subset. These calculations used the full unit-normalized embeddings without coordinate-wise min-max scaling.

\begin{readerplot}
% Generated by analysis/export_tikz_plots.py; edit the exporter.
\begin{tikzpicture}
\begin{axis}[arborea axis,xmin=1,xmax=167,xtick={1,50,100,150},xlabel={Component rank},title={(a) Individual components},legend columns=2,arborea legend below,ymode=log,ymin=0.000005,ymax=.8,ylabel={Variance share}]
\addplot[arbHuman,solid,no marks,arborea data] table[x=x,y=mean] {figures/tikz/data/spectrum_variance_fraction_baseline.dat};
\addplot[draw=none,no marks,name path=lononce,forget plot] table[x=x,y=lo] {figures/tikz/data/spectrum_variance_fraction_nonce.dat};
\addplot[draw=none,no marks,name path=hinonce,forget plot] table[x=x,y=hi] {figures/tikz/data/spectrum_variance_fraction_nonce.dat};
\addplot[arbOrange,fill opacity=.15,draw=none,forget plot] fill between[of=lononce and hinonce];
\addplot[arbOrange,densely dotted,no marks,arborea data] table[x=x,y=mean] {figures/tikz/data/spectrum_variance_fraction_nonce.dat};
\addplot[draw=none,no marks,name path=loprior_titles,forget plot] table[x=x,y=lo] {figures/tikz/data/spectrum_variance_fraction_prior_titles.dat};
\addplot[draw=none,no marks,name path=hiprior_titles,forget plot] table[x=x,y=hi] {figures/tikz/data/spectrum_variance_fraction_prior_titles.dat};
\addplot[arbBlue,fill opacity=.15,draw=none,forget plot] fill between[of=loprior_titles and hiprior_titles];
\addplot[arbBlue,dashed,no marks,arborea data] table[x=x,y=mean] {figures/tikz/data/spectrum_variance_fraction_prior_titles.dat};
\addplot[draw=none,no marks,name path=lohierarchical,forget plot] table[x=x,y=lo] {figures/tikz/data/spectrum_variance_fraction_hierarchical.dat};
\addplot[draw=none,no marks,name path=hihierarchical,forget plot] table[x=x,y=hi] {figures/tikz/data/spectrum_variance_fraction_hierarchical.dat};
\addplot[arbGreen,fill opacity=.15,draw=none,forget plot] fill between[of=lohierarchical and hihierarchical];
\addplot[arbGreen,dashdotted,no marks,arborea data] table[x=x,y=mean] {figures/tikz/data/spectrum_variance_fraction_hierarchical.dat};
\end{axis}
\end{tikzpicture}
\par\smallskip
\begin{tikzpicture}
\begin{axis}[arborea axis,xmin=1,xmax=167,xtick={1,50,100,150},xlabel={Component rank},title={(b) Accumulated variance},legend columns=2,arborea legend below,ymin=0,ymax=1.02,ytick={0,.5,1},ylabel={Cumulative share}]
\addplot[arbHuman,solid,no marks,arborea data] table[x=x,y=mean] {figures/tikz/data/spectrum_cumulative_fraction_baseline.dat};
\addlegendentry{Human}
\addplot[draw=none,no marks,name path=lononce,forget plot] table[x=x,y=lo] {figures/tikz/data/spectrum_cumulative_fraction_nonce.dat};
\addplot[draw=none,no marks,name path=hinonce,forget plot] table[x=x,y=hi] {figures/tikz/data/spectrum_cumulative_fraction_nonce.dat};
\addplot[arbOrange,fill opacity=.15,draw=none,forget plot] fill between[of=lononce and hinonce];
\addplot[arbOrange,densely dotted,no marks,arborea data] table[x=x,y=mean] {figures/tikz/data/spectrum_cumulative_fraction_nonce.dat};
\addlegendentry{Nonce}
\addplot[draw=none,no marks,name path=loprior_titles,forget plot] table[x=x,y=lo] {figures/tikz/data/spectrum_cumulative_fraction_prior_titles.dat};
\addplot[draw=none,no marks,name path=hiprior_titles,forget plot] table[x=x,y=hi] {figures/tikz/data/spectrum_cumulative_fraction_prior_titles.dat};
\addplot[arbBlue,fill opacity=.15,draw=none,forget plot] fill between[of=loprior_titles and hiprior_titles];
\addplot[arbBlue,dashed,no marks,arborea data] table[x=x,y=mean] {figures/tikz/data/spectrum_cumulative_fraction_prior_titles.dat};
\addlegendentry{Prior titles}
\addplot[draw=none,no marks,name path=lohierarchical,forget plot] table[x=x,y=lo] {figures/tikz/data/spectrum_cumulative_fraction_hierarchical.dat};
\addplot[draw=none,no marks,name path=hihierarchical,forget plot] table[x=x,y=hi] {figures/tikz/data/spectrum_cumulative_fraction_hierarchical.dat};
\addplot[arbGreen,fill opacity=.15,draw=none,forget plot] fill between[of=lohierarchical and hihierarchical];
\addplot[arbGreen,dashdotted,no marks,arborea data] table[x=x,y=mean] {figures/tikz/data/spectrum_cumulative_fraction_hierarchical.dat};
\addlegendentry{Concept-guided}
\addplot[black!45,densely dotted,forget plot] coordinates {(1,.9) (167,.9)};
\end{axis}
\end{tikzpicture}\par
\caption{Distribution of variation across embedding directions at matched sample size, $n=168$.}
\label{fig:matched-spectra}
\end{readerplot}

A steep decline in the individual variance shares, accompanied by a rapid rise in cumulative variance, indicates that variation is concentrated in relatively few directions. A higher participation ratio indicates that more directions contribute substantially. This remains a geometric proxy: it does not establish semantic validity, count distinct cuisines, or measure the total amount of variation.

\begin{table}[!htbp]
\captionsetup{type=table}
\phantomcaption
\label{tab:participation-ratio}
\small\setlength{\tabcolsep}{3pt}
\begin{tabularx}{\linewidth}{@{}Xrrr@{}}\toprule
Population & \shortstack{Participation\\ratio} & \shortstack{PR SD\\(subsets)} & \shortstack{Top-ten\\variance (\%)} \\\midrule
\textit{Human reference} & \textit{53.70} & \textit{---} & \textit{34.3} \\
Concept-guided & \textbf{48.85} & 1.19 & \textbf{36.5} \\
Prior titles & 25.86 & 1.01 & 52.7 \\
Nonce & 4.37 & 0.30 & 93.8 \\
\bottomrule\end{tabularx}
\end{table}

Concept-guided prompting had the highest participation ratio among the generated populations and was nearest to the human reference. Its variation was therefore distributed across more effective embedding directions than either baseline, while remaining somewhat more concentrated than human variation.

The first ten components accounted for 93.8\% of nonce variance, compared with 52.7\% for prior titles and 36.5\% for concept-guided prompting. The corresponding human value was 34.3\%. Both summaries therefore support the same finding: concept-guided variation was less concentrated than either baseline and closer to the human reference in how variance was distributed.

Each population's spectrum uses its own principal directions. Similar spectra consequently do not establish that populations vary along the same directions or occupy the same locations in embedding space.

\subsection{Comparative Centroid Alignment}
\label{sec:centroid-alignment}

Centroid alignment measures whether two recipe populations have a similar average semantic character. Imagine describing recipes using spiciness and sweetness. A population containing mostly spicy dishes would have an average arrow pointing toward spiciness, whereas a population containing mostly sweet dishes would point toward sweetness. A smaller angle between two populations' average arrows indicates a more similar balance of features (Figure~\ref{fig:centroid-alignment}). These named features are teaching examples; actual embeddings contain many learned dimensions.

For the calculation, each recipe's embedding vector was normalized to unit length. We averaged these vectors within each population to obtain its centroid, then normalized the centroid to retain its direction. The cosine distance between each generated population's centroid direction and the human-reference centroid direction was calculated as:
\begin{equation}
d_{\mathrm{centroid}}=1-\cos(\theta),
\end{equation}
where $\theta$ is the angle between the two centroid directions in the full embedding space. Smaller distances indicate closer average alignment; zero indicates identical directions. All 1,000 recipes in each generated population and all 168 human recipes contributed to their respective centroids.

\begin{readerplot}
\pgfplotsset{title={Centroid alignment}}
% Methods TikZ base graph with measured recipe data.
\begin{tikzpicture}[every node/.style={inner sep=1pt},line cap=round]
\begin{axis}[arborea scatter,xmin=0,xmax=1.12,ymin=0,ymax=1.12,
xtick={0,.25,.5,.75,1},ytick={0,.25,.5,.75,1},
xlabel={Projection axis 1},ylabel={Projection axis 2},clip=false,
legend columns=2,arborea legend below]
\draw[-{Stealth[length=3pt]},arbBlue,arborea data] (0,0) -- (0.519661454394,0.793354348143);
\draw[arbBlue] plot[only marks,mark=triangle,mark size=\arbMarkerSize,mark options={fill=white,arborea marker edge}] coordinates {(0.350356726998,0.664304223242) (0.505008597873,0.323803962982) (0.387734571836,0.504772592062) (0.365278950148,0.622001829308) (0.389832186671,0.629585231014) (0.348336093994,0.647925160245) (0.43900596397,0.473055395314) (0.40356753128,0.525552558509) (0.406132603691,0.614435849841) (0.386935455245,0.610984810343) (0.387193632758,0.603914508919) (0.402376373491,0.651530496795)};
\draw[-{Stealth[length=3pt]},arbOrange,arborea data] (0,0) -- (0.454054795425,0.823590263011);
\draw[arbOrange] plot[only marks,mark=diamond,mark size=\arbMarkerSize,mark options={fill=white,arborea marker edge}] coordinates {(0.362289116749,0.680110774045) (0.364429014835,0.676696317928) (0.360963969935,0.689255097733) (0.358323007572,0.734416638413) (0.367227079307,0.679725696642) (0.374100834388,0.697102674051) (0.345208895758,0.708618290003) (0.359765534057,0.691271139616) (0.35332997266,0.727324370064) (0.368224647557,0.68319916583) (0.357954824724,0.726773115937) (0.386737249826,0.691939471547)};
\draw[-{Stealth[length=3pt]},arbGreen,arborea data] (0,0) -- (0.589564528221,0.669150098811);
\draw[arbGreen] plot[only marks,mark=o,mark size=\arbMarkerSize,mark options={fill=white,arborea marker edge}] coordinates {(0.403178929653,0.346830178748) (0.349911307746,0.455835285969) (0.380628704236,0.457752665702) (0.402384446949,0.443753989834) (0.381433672698,0.333956683376) (0.425257076932,0.385740731227) (0.434535134079,0.413298967808) (0.357519616731,0.496075510356) (0.424477524904,0.522918387829) (0.407391910753,0.384263504768) (0.413830206726,0.414335283521) (0.367132877943,0.39788695564)};
\draw[-{Stealth[length=3pt]},arbHuman,arborea data,dashed] (0,0) -- (0.965925826289,0.258819045103);
\draw[arbHuman] plot[only marks,mark=square,mark size=\arbMarkerSize,mark options={fill=white,arborea marker edge}] coordinates {(0.55916503103,0.188369196518) (0.585552008188,0.108310192711) (0.461157990032,0.189547515612) (0.597884925238,0.0526478704006) (0.358880244614,0.299778141564) (0.535672158785,0.15806628102) (0.538298442748,0.220114361702) (0.593397897623,0.201918465212) (0.602922568786,0.0532822865841) (0.589077197396,0.12241486444) (0.466096064969,0.147231731971) (0.59077581932,0.162889060683)};
\addplot[arbGreen,no marks,arborea data,domain=15:48.617861660116965,samples=30,forget plot] ({0.24*cos(x)},{0.24*sin(x)});
\node[arbGreen,rotate=31.808930830058483,fill=white] at (0.263441268889126,0.1633973618088346) {$33.6^\circ$};
\addplot[arbBlue,no marks,arborea data,domain=15:56.77447566715517,samples=30,forget plot] ({0.42*cos(x)},{0.42*sin(x)});
\node[arbBlue,rotate=35.887237833577586,fill=white] at (0.3969843926991114,0.28723403689903765) {$41.8^\circ$};
\addplot[arbOrange,no marks,arborea data,domain=15:61.13154587958992,samples=30,forget plot] ({0.6*cos(x)},{0.6*sin(x)});
\node[arbOrange,rotate=38.06577293979496,fill=white] at (0.5274933337897566,0.4130989987973444) {$46.1^\circ$};
\node[arbHuman] at (.84,.025) {Human average};
\addlegendimage{arbBlue,only marks,mark=triangle}
\addlegendentry{Prior titles}
\addlegendimage{arbOrange,only marks,mark=diamond}
\addlegendentry{Nonce}
\addlegendimage{arbGreen,only marks,mark=o}
\addlegendentry{Concept-guided}
\addlegendimage{arbHuman,only marks,mark=square}
\addlegendentry{Human}
\end{axis}
\end{tikzpicture}\par
\caption{Centroid alignment using measured recipe embeddings.}
\label{fig:centroid-alignment}
\end{readerplot}

\begin{table}[!htbp]
\captionsetup{type=table,position=bottom}
\small\setlength{\tabcolsep}{3pt}
\begin{tabularx}{\linewidth}{@{}Xr@{}}\toprule
Population & \shortstack{Centroid cosine distance\\from human reference} \\\midrule
\textit{Human reference} & \textit{0.000} \\
Concept-guided & \textbf{0.257} \\
Prior titles & 0.293 \\
Nonce & 0.348 \\
\bottomrule\end{tabularx}
\caption{Full-dimensional distances between normalized centroid directions. The human reference has zero distance from itself.}
\label{tab:centroid-alignment}
\end{table}

Concept-guided prompting had the smallest centroid distance from the human reference, followed by prior titles and nonce. Its population therefore had the closest average embedding direction to the human population among the three generated strategies. The distance remained nonzero, indicating that their average directions did not coincide.

A separate supporting calculation measured within-population spread by averaging cosine distances over all distinct pairs of recipes within each population. This measures internal dispersion rather than centroid alignment and is calculated from the full embeddings, rather than the displayed sample.

\begin{table}[!htbp]
\captionsetup{type=table,position=bottom}
\small\setlength{\tabcolsep}{3pt}
\begin{tabularx}{\linewidth}{@{}Xr@{}}\toprule
Population & \shortstack{Mean pairwise\\cosine distance} \\\midrule
\textit{Human reference} & \textit{0.671} \\
Concept-guided & \textbf{0.564} \\
Prior titles & 0.441 \\
Nonce & 0.335 \\
\bottomrule\end{tabularx}
\caption{Within-population spread, measured separately from centroid alignment using all distinct recipe pairs.}
\label{tab:pairwise-spread}
\end{table}

Concept-guided prompting also had the greatest pairwise spread among the generated populations and was nearest to the human value, while remaining below it. Together, these two measurements show a closer average direction and a dispersion closer to the human reference than either baseline. They do not establish that the full distributions coincide or that individual concept-guided recipes are closer to human recipes.

\subsection{Comparative Generated-to-Human Proximity}
\label{sec:generated-to-human}

This analysis was made to understand how closely individual generated recipes resembled examples from the human population. In other words, for each generated recipe, how close was the most similar human recipe available?

To determine this, each recipe's embedding vector was normalized to unit length. We then calculated the cosine distance between each generated recipe and all 168 human recipes, retaining the smallest distance. For unit-normalized vectors $g$ and $h$, cosine distance is:
\begin{equation}
d_{\cos}(g,h)=1-g^\top h.
\end{equation}
The mean of these nearest distances provides the generated population's proximity to individual human examples:
\begin{equation}
D_{G\rightarrow H}=\frac{1}{|G|}\sum_{g\in G}\min_{h\in H}d_{\cos}(g,h),
\end{equation}
where $G$ is a generated population and $H$ is the human reference population. Each strategy was evaluated using all 1,000 generated recipes against the same 168 human recipes. Each generated recipe contributes one distance, and the same human recipe can be the nearest example for multiple generated recipes.

\begin{readerplot}
% Generated by analysis/export_reference_proximity.py; cached recipe embeddings.
\begin{tikzpicture}\begin{axis}[arborea axis,xmin=0,xmax=1,xtick={0,.25,.5,.75,1},ymin=0,ymax=9.0,title={Generated-to-human proximity},xlabel={Nearest cosine distance},ylabel={Density},legend columns=2,arborea legend below]
\addplot[arbBlue,dashed,arborea density,fill=arbBlue] table[x=x,y=y] {figures/tikz/data/results_generated_to_human_kde_prior_titles.dat}\closedcycle;
\addlegendentry{Prior titles}
\addplot[arbBlue,dashed,no marks,forget plot] coordinates {(0.4546622469092069,0) (0.4546622469092069,9.0)};
\addplot[arbOrange,densely dotted,arborea density,fill=arbOrange] table[x=x,y=y] {figures/tikz/data/results_generated_to_human_kde_nonce.dat}\closedcycle;
\addlegendentry{Nonce}
\addplot[arbOrange,dashed,no marks,forget plot] coordinates {(0.38547079418937186,0) (0.38547079418937186,9.0)};
\addplot[arbGreen,dashdotted,arborea density,fill=arbGreen] table[x=x,y=y] {figures/tikz/data/results_generated_to_human_kde_hierarchical.dat}\closedcycle;
\addlegendentry{Concept-guided}
\addplot[arbGreen,dashed,no marks,forget plot] coordinates {(0.4998097619251651,0) (0.4998097619251651,9.0)};
\end{axis}\end{tikzpicture}\par
\phantomcaption
\label{fig:generated-to-human}
\end{readerplot}

Smaller values indicate that generated recipes lie closer to available human examples in the embedding space. This measures proximity to individual examples rather than similarity between population averages. It remains a geometric proxy and does not establish recipe quality.

\begin{table}[!htbp]
\captionsetup{type=table,position=bottom}
\small\setlength{\tabcolsep}{3pt}
\begin{tabularx}{\linewidth}{@{}Xr@{}}\toprule
\shortstack[l]{Generated\\population} & \shortstack{Mean nearest-\\human distance} \\\midrule
Concept-guided & 0.500 \\
Prior titles & 0.455 \\
Nonce & \textbf{0.385} \\
\bottomrule\end{tabularx}
\caption{Mean generated-to-human cosine distances. Smaller values indicate closer proximity to individual human examples.}
\label{tab:generated-to-human}
\end{table}

Nonce had the smallest mean distance, followed by prior titles and concept-guided prompting. The average concept-guided recipe was therefore farther from its nearest human example than the average recipe from either baseline.

This finding differs from the centroid comparison. Concept-guided prompting had the closest average population direction to the human reference, but its individual outputs were farther from their nearest human examples on average. These measurements concern different properties: a similar average direction does not require individual recipes to have close human counterparts.

A small generated-to-human distance also does not establish that a strategy approaches many different human recipes. Multiple generated recipes may be close to the same human example. To examine proximity across the human collection, the next analysis reverses the starting population.

\subsection{Comparative Human-to-Generated Proximity}
\label{sec:human-to-generated}

This analysis was made to understand how closely each strategy approached the examples represented in the human population. In other words, starting from each human recipe, how close was the most similar recipe produced by each strategy?

For each of the 168 human recipes, we calculated the cosine distance to all 1,000 recipes in a generated population and retained the smallest distance. We repeated this calculation separately for each strategy, then averaged the distances across the human population:
\begin{equation}
D_{H\rightarrow G}=\frac{1}{|H|}\sum_{h\in H}\min_{g\in G}d_{\cos}(h,g).
\end{equation}
Each human recipe contributes one distance per strategy. Although the cosine distance between any particular pair of recipes is unchanged when the pair is reversed, the two population-level measures differ because they select and average nearest neighbors from different starting populations.

\begin{readerplot}
% Generated by analysis/export_reference_proximity.py; cached recipe embeddings.
\begin{tikzpicture}\begin{axis}[arborea axis,xmin=0,xmax=1,xtick={0,.25,.5,.75,1},ymin=0,ymax=9.0,title={Human-to-generated proximity},xlabel={Nearest cosine distance},ylabel={Density},legend columns=2,arborea legend below]
\addplot[arbBlue,dashed,arborea density,fill=arbBlue] table[x=x,y=y] {figures/tikz/data/results_human_to_generated_kde_prior_titles.dat}\closedcycle;
\addlegendentry{Prior titles}
\addplot[arbBlue,dashed,no marks,forget plot] coordinates {(0.5094692256206538,0) (0.5094692256206538,9.0)};
\addplot[arbOrange,densely dotted,arborea density,fill=arbOrange] table[x=x,y=y] {figures/tikz/data/results_human_to_generated_kde_nonce.dat}\closedcycle;
\addlegendentry{Nonce}
\addplot[arbOrange,dashed,no marks,forget plot] coordinates {(0.5707994191059875,0) (0.5707994191059875,9.0)};
\addplot[arbGreen,dashdotted,arborea density,fill=arbGreen] table[x=x,y=y] {figures/tikz/data/results_human_to_generated_kde_hierarchical.dat}\closedcycle;
\addlegendentry{Concept-guided}
\addplot[arbGreen,dashed,no marks,forget plot] coordinates {(0.4723249213329236,0) (0.4723249213329236,9.0)};
\end{axis}\end{tikzpicture}\par
\phantomcaption
\label{fig:human-to-generated}
\end{readerplot}

Smaller values indicate that the generated population approaches the human examples more closely on average. Each human recipe receives equal weight, including those that are distant from every generated recipe. The measure describes proximity across this reference collection; it does not establish that every human example has a sufficiently close generated counterpart.

\begin{table}[!htbp]
\captionsetup{type=table,position=bottom}
\small\setlength{\tabcolsep}{3pt}
\begin{tabularx}{\linewidth}{@{}Xr@{}}\toprule
\shortstack[l]{Generated\\population} & \shortstack{Mean nearest-\\generated distance} \\\midrule
Concept-guided & \textbf{0.472} \\
Prior titles & 0.509 \\
Nonce & 0.571 \\
\bottomrule\end{tabularx}
\caption{Mean human-to-generated cosine distances. Each human recipe is compared against 1,000 generated recipes per strategy. Smaller values indicate closer proximity across human reference queries.}
\label{tab:human-to-generated}
\end{table}

Concept-guided prompting had the smallest mean distance, followed by prior titles and nonce. Starting from the human recipes, the nearest concept-guided recipe was therefore closer on average than the nearest recipe from either baseline.

This reverses the ordering of the generated-to-human comparison. The two findings are compatible: a generated population can approach the human collection more closely across human queries while its average output remains farther from its nearest human example.

Together, the comparisons show that concept-guided prompting approached the available human references more closely on average, but did not produce the closest individual outputs on average. The next analysis asks how many human references were reached within a locally defined distance threshold.

\subsection{Comparative Human-Reference Coverage}
\label{sec:human-reference-coverage}

This analysis was made to understand how many neighborhoods in the human recipe population were reached by each generated population. In other words, how many human recipes had at least one generated recipe sufficiently close to them?

To determine this, we defined a neighborhood around each of the 168 human recipes. Its radius was the cosine distance to its fifth-nearest other human recipe, excluding the starting recipe itself. Each human recipe therefore had its own threshold for what counted as nearby.

For a human recipe $h$, this radius is:
\begin{equation}
r_h=d_{\cos}\bigl(h,h_{(5)}\bigr),
\end{equation}
where $h_{(5)}$ is its fifth-nearest other human recipe. A neighborhood was counted as covered when at least one generated recipe lay within or on its boundary:
\begin{equation}
C(G)=\frac{1}{|H|}\sum_{h\in H}\mathbf{1}\left[\min_{g\in G}d_{\cos}(h,g)\leq r_h\right].
\end{equation}
Here, $\mathbf{1}$ equals 1 when the condition is satisfied and 0 otherwise. Multiplying $C(G)$ by 100 gives the percentage of human neighborhoods covered.

For example, if a human recipe's fifth human neighbor is at cosine distance 0.30, a generated recipe at distance 0.25 covers that neighborhood. If the nearest generated recipe is at distance 0.35, the neighborhood remains uncovered. Multiple generated recipes within the same neighborhood still contribute only one covered neighborhood.

\begin{readerplot}
\pgfplotsset{title={Human-reference coverage}}
% Generated by analysis/export_results_method_plots.py; measured recipe data.
\begin{tikzpicture}
\begin{axis}[arborea axis,arborea scatter,xmin=-0.554365137084025,xmax=0.45287502793590817,ymin=-0.544974414883602,ymax=0.4622657501363312,xlabel={PC 1},ylabel={PC 2},legend columns=2,arborea legend below]
\addplot[only marks,mark=triangle*,arborea highlight,arbBlue,mark options={draw=white,arborea marker edge}] table[x=x,y=y] {figures/tikz/data/results_coverage_prior_titles.dat};
\addlegendentry{Prior titles}
\addplot[only marks,mark=triangle*,arborea highlight,arbOrange,mark options={draw=white,arborea marker edge}] table[x=x,y=y] {figures/tikz/data/results_coverage_nonce.dat};
\addlegendentry{Nonce}
\addplot[only marks,mark=triangle*,arborea highlight,arbCoral,mark options={draw=white,arborea marker edge}] table[x=x,y=y] {figures/tikz/data/results_coverage_hierarchical.dat};
\addlegendentry{Concept-guided}
\draw[arbBlue,line width=0.8pt] (axis cs:0.19907924289527246,0.09828290210847919) circle[radius=3.9pt];
\draw[arbCoral,line width=0.8pt] (axis cs:0.19907924289527246,0.09828290210847919) circle[radius=3.1pt];
\draw[arbBlue,line width=0.5333333333333333pt] (axis cs:-0.3143266740084163,0.1749298324323343) circle[radius=4.033333333333333pt];
\draw[arbOrange,line width=0.5333333333333333pt] (axis cs:-0.3143266740084163,0.1749298324323343) circle[radius=3.5pt];
\draw[arbCoral,line width=0.5333333333333333pt] (axis cs:-0.3143266740084163,0.1749298324323343) circle[radius=2.966666666666667pt];
\draw[arbBlue,line width=0.8pt] (axis cs:0.005292206619039144,0.24122090621785297) circle[radius=3.9pt];
\draw[arbCoral,line width=0.8pt] (axis cs:0.005292206619039144,0.24122090621785297) circle[radius=3.1pt];
\draw[arbBlue,line width=0.5333333333333333pt] (axis cs:0.14293788147153805,-0.11367453891771798) circle[radius=4.033333333333333pt];
\draw[arbOrange,line width=0.5333333333333333pt] (axis cs:0.14293788147153805,-0.11367453891771798) circle[radius=3.5pt];
\draw[arbCoral,line width=0.5333333333333333pt] (axis cs:0.14293788147153805,-0.11367453891771798) circle[radius=2.966666666666667pt];
\draw[arbBlue,line width=0.8pt] (axis cs:-0.13107727514168396,0.11413313447189423) circle[radius=3.9pt];
\draw[arbCoral,line width=0.8pt] (axis cs:-0.13107727514168396,0.11413313447189423) circle[radius=3.1pt];
\draw[arbBlue,line width=0.8pt] (axis cs:0.32057739743149644,-0.02074407588687) circle[radius=3.9pt];
\draw[arbCoral,line width=0.8pt] (axis cs:0.32057739743149644,-0.02074407588687) circle[radius=3.1pt];
\draw[arbBlue,line width=1.6pt] (axis cs:0.06210848368254924,0.10801135121868449) circle[radius=3.5pt];
\draw[arbBlue,line width=0.5333333333333333pt] (axis cs:0.04079814796861541,-0.039652432308468766) circle[radius=4.033333333333333pt];
\draw[arbOrange,line width=0.5333333333333333pt] (axis cs:0.04079814796861541,-0.039652432308468766) circle[radius=3.5pt];
\draw[arbCoral,line width=0.5333333333333333pt] (axis cs:0.04079814796861541,-0.039652432308468766) circle[radius=2.966666666666667pt];
\draw[arbBlue,line width=0.8pt] (axis cs:0.31678910160998697,0.1365391063059413) circle[radius=3.9pt];
\draw[arbCoral,line width=0.8pt] (axis cs:0.31678910160998697,0.1365391063059413) circle[radius=3.1pt];
\draw[arbInk!25,line width=1.6pt] (axis cs:0.12043573486557284,0.22724968678447904) circle[radius=3.5pt];
\draw[arbInk!25,line width=1.6pt] (axis cs:0.20823699950678098,-0.02764289928752541) circle[radius=3.5pt];
\draw[arbBlue,line width=1.6pt] (axis cs:0.12115659675229025,0.024870340678752857) circle[radius=3.5pt];
\draw[arbInk!25,line width=1.6pt] (axis cs:0.21446134158741895,0.20314886399193807) circle[radius=3.5pt];
\draw[arbInk!25,line width=1.6pt] (axis cs:0.28696692074359514,0.06021840441529769) circle[radius=3.5pt];
\draw[arbBlue,line width=0.8pt] (axis cs:0.2788409818507226,-0.07881485530260662) circle[radius=3.9pt];
\draw[arbCoral,line width=0.8pt] (axis cs:0.2788409818507226,-0.07881485530260662) circle[radius=3.1pt];
\draw[arbInk!25,line width=1.6pt] (axis cs:0.12258024783528745,0.1566361425659128) circle[radius=3.5pt];
\draw[arbInk!25,line width=1.6pt] (axis cs:0.03688770546853066,0.1686359468999163) circle[radius=3.5pt];
\draw[arbInk!25,line width=1.6pt] (axis cs:0.24061700435296232,0.1450028231433877) circle[radius=3.5pt];
\draw[arbBlue,line width=0.8pt] (axis cs:0.18260155497344202,0.029665022786124175) circle[radius=3.9pt];
\draw[arbCoral,line width=0.8pt] (axis cs:0.18260155497344202,0.029665022786124175) circle[radius=3.1pt];
\draw[arbBlue,line width=0.8pt] (axis cs:0.12497742008057733,-0.036553706699856595) circle[radius=3.9pt];
\draw[arbCoral,line width=0.8pt] (axis cs:0.12497742008057733,-0.036553706699856595) circle[radius=3.1pt];
\draw[arbCoral,line width=1.6pt] (axis cs:0.14081161399854408,0.0863328742940077) circle[radius=3.5pt];
\draw[arbInk!25,line width=1.6pt] (axis cs:0.06207500489505465,0.22136354155712668) circle[radius=3.5pt];
\draw[arbBlue,line width=0.5333333333333333pt] (axis cs:0.24097716653991774,0.026936095785091485) circle[radius=4.033333333333333pt];
\draw[arbOrange,line width=0.5333333333333333pt] (axis cs:0.24097716653991774,0.026936095785091485) circle[radius=3.5pt];
\draw[arbCoral,line width=0.5333333333333333pt] (axis cs:0.24097716653991774,0.026936095785091485) circle[radius=2.966666666666667pt];
\draw[arbInk!25,line width=1.6pt] (axis cs:0.17262234622291894,0.1712732900379263) circle[radius=3.5pt];
\addplot[only marks,mark=square*,arbInk!65,mark options={draw=white,arborea marker edge}] table[x=x,y=y] {figures/tikz/data/results_coverage_human.dat};
\addlegendentry{Human reference}
\end{axis}
\end{tikzpicture}\par
\caption{Coverage of human-reference neighborhoods using measured recipe embeddings.}
\label{fig:human-reference-coverage}
\end{readerplot}

Higher coverage indicates that more human-reference neighborhoods contain at least one generated recipe. Neighborhoods may overlap, and one generated recipe may cover several human references. Coverage therefore counts reference neighborhoods reached rather than distinct recipe categories or non-overlapping regions.

\begin{table}[!htbp]
\captionsetup{type=table,position=bottom}
\small\setlength{\tabcolsep}{3pt}
\begin{tabularx}{\linewidth}{@{}Xrr@{}}\toprule
\shortstack[l]{Generated\\population} & \shortstack{Human neighborhoods\\covered} & Coverage \\\midrule
Concept-guided & \textbf{79 of 168} & \textbf{47.0\%} \\
Prior titles & 59 of 168 & 35.1\% \\
Nonce & 22 of 168 & 13.1\% \\
\bottomrule\end{tabularx}
\caption{Human-reference coverage using each human recipe's fifth-other-human cosine distance as its neighborhood radius.}
\label{tab:human-reference-coverage}
\end{table}

Concept-guided prompting reached the largest number of human neighborhoods: 20 more than prior titles and 57 more than nonce. This supports broader coverage of the available human reference collection under the specified neighborhood rule.

However, concept-guided coverage remained below half of the human neighborhoods. Its greater coverage therefore represents an improvement relative to the two baselines, rather than comprehensive representation of the human collection.

The coverage result complements the lower human-to-generated mean distance: concept-guided recipes were closer on average across human queries, and more human neighborhoods contained a concept-guided recipe. Neither finding establishes recipe quality or coverage beyond the available human corpus.

\section{Discussion and conclusion}

From the interpretation of the results, concept-guided prompting produced a broader population of recipes that more closely resembled certain aspects of the human-authored corpus. It also reached more embedded vector neighborhoods. Thus, it is clear that using a concept taxonomy can have benefits when it comes to producing diverse sets of solutions and generated texts. Although the figures and results are positive in nearly every case for the concept-guided populations, because we are using embeddings as a proxy for semantic content, we have to consider further testing of the quality and quantitative aspects of generated content using concept-guided prompting in later studies.

Furthermore, comparison against a human corpus shackled the experiment to 'useful diversity.' There could, theoretically, exist strategies that are great at producing diverse outputs that are nothing like human-authored recipes and are therefore completely useless to the task at hand. Using multiple analyses and studies helped to identify that the internal diversity was useful.

As to why this strategy works well, a plausible explanation is that generating thematic concepts first distributes requests into semantic islands that then steer a model toward outputs in a certain constrained response space. Ergo, it may perhaps be the most useful to generate the widest semantic range of concepts first to then distribute responses into. A less diverse set of concepts may lead to a less diverse set of solutions. This still has to be studied further.

This strategy aligns well with prior work that organizes content before generating text for a given task. Shao et al.\ first plan groups of input items and then generate sentences conditioned on that plan \citep{shao2019phvm}. Young et al.\ use chain-of-specification prompting to generate structural specifications before producing text that satisfies them \citep{young2024cos}. However, our study examines this similar approach at the population level rather than individually or linearly in the case of task execution like automated code generation.

Prior work also establishes the importance of recognizing the distinction between lexical and embedding results. Han et al.\ found that lexical diversity measures can miss semantic variation in dialogue \citep{han2022semanticdiversity}; Zhang et al.\ found that form-based metrics can overestimate diversity in commonsense generation \citep{zhang2025diversityeval}.

For systems that generate candidates for later evaluation, such as LLM-guided program search, the findings motivate testing concept-guided prompting as a way to broaden the pool presented to an evaluator. This connects to the population-level objective of MAP-Elites, which retains high-performing solutions  across feature regions \citep{mouret2015mapelites}, and QDAIF, which uses language-model feedback to evaluate quality and diversity during evolutionary search \citep{bradley2023qdaif}. Our procedure addresses candidate generation without quality-guided selection or iterative evaluator feedback. Whether its broader pool yields better selected recipes, more robust solutions, or lower total search cost remains a downstream question. Broader embedding coverage alone cannot establish those benefits, as an embedding is not a quantitative score of quality.

Concept-guided prompting therefore offers a concrete generation design for further study: distribute a collection across concepts before generating its members. In the saved comparison, this design combined broader internal variation with closer agreement on selected human-population summaries and greater reference coverage. Its value as a source of better solutions remains to be established through independent runs and quality evaluation.

\bigskip
\begingroup
\footnotesize
\raggedright
\setlength{\bibsep}{0.5pt}
\bibliographystyle{plainnat}
\bibliography{ref}
\endgroup

\end{document}